%% file: main.tex
\documentclass{article}

    \PassOptionsToPackage{numbers, compress}{natbib}

\usepackage[final]{neurips_2021}

\usepackage{algorithmic}
\usepackage[ruled, lined, linesnumbered, commentsnumbered, longend]{algorithm2e}
\usepackage[utf8]{inputenc} 
\usepackage[T1]{fontenc}    
\usepackage{booktabs}       
\usepackage{amsfonts}       
\usepackage{nicefrac}       
\usepackage{microtype}      
\usepackage{xcolor}      
\usepackage{graphicx}   
\usepackage{subfigure}
\input{math.tex}

\title{Debugging using Orthogonal Gradient Descent}

\author{%
 Narsimha Chilkuri \\
Independent Researcher\\
  \texttt{nrchilku@uwaterloo.ca}
  \And
  Chris Eliasmith\\
  University of Waterloo \\
  \texttt{celiasmith@uwaterloo.ca} \\
}

\begin{document}

\maketitle

\begin{abstract}
In this report we consider the following problem: Given a trained model that is partially faulty, can we correct its behaviour without having to train the model from scratch? In other words, can we ``debug" neural networks similar to how we address bugs in our mathematical models and standard computer code. We base our approach on the hypothesis that  debugging can be treated as a two-task continual learning problem. In particular, we employ a modified version of a continual learning algorithm called Orthogonal Gradient Descent (OGD) to demonstrate, via two simple experiments on the MNIST dataset, that we can in-fact \textit{unlearn} the undesirable behaviour while retaining the general performance of the model, and we can additionally \textit{relearn} the appropriate behaviour, both without having to train the model from scratch.
\end{abstract}

\section{Introduction}
While the field of machine learning has seen incredible progress in the past decade, in our view, there is an important problem that has not received enough traction: the problem of ``debugging" machine learning models. That is, when it is discovered that a trained model is partially faulty -- that it responds to a portion of the inputs in an incorrect manner, perhaps due to a dataset bias or due to insufficient training data -- a very common method of debugging is to simply train the model from scratch. While this naive approach works well, retraining models can be prohibitively expensive, with many of the state-of-the art models requiring several weeks and multiple GPUs to train. Finances and the time overhead aside, as a matter of principle, starting from scratch is unsatisfactory. For instance, we rarely address bugs in our mathematical models or standard computer code by throwing everything away and starting with a clean slate -- and neither do we abandon a child every time they misbehave or say unpleasant things!

With the above concerns in mind, we explore the problem of debugging machine learning models without having to train the models from scratch. We do so by basing our approach on the claim that the  {\it debugging problem can be treated as a two-task continual learning problem}. More specifically, we employ a modified version of a continual learning method known as Orthogonal Gradient Descent \citep{farajtabar2020orthogonal}, and using a simple convolutional neural network trained on a (purposefully) corrupted version of the MNIST dataset we show that (1) it is possible to unlearn the way the model responds to the misclassified data points without affecting the general performance; and (2) on top of unlearning, it is also possible to relearn the right responses to misclassified inputs and attain the maximum possible test accuracy (with the simple network), without training from scratch.

\section{Background}



In this section we talk about Orthogonal Gradient Descent (OGD), the method that underlies our approach. While continual learning, in general, focuses on the problem of sequential learning of many tasks  $\{\mathcal{T}_1, \mathcal{T}_2,..., \mathcal{T}_n\}$, here we discuss the method with just two-tasks in mind, $\{\mathcal{T}_1, \mathcal{T}_2\}$. We use $f(x; \theta)$ to denote the neural network, where $x \in \R^d$ is the input to the network and $\theta$ are the weights that parametrize the model. The network takes in an input $x$ and outputs its prediction $y \in \R^c$. In case of classification problems, $f_j(x; \theta)$ denotes the $j$-th logit associated to the $j$-th class. The loss function (for a single task) is denoted as 
\begin{align*}
    \mathcal{L}(\theta) = \sum_i l(\tilde{y}_i, f(x_i; \theta)),
\end{align*}
where the sum is over the training data and $\tilde{y}$ denotes the ground truth.

When discussing OGD, it is crucial to distinguish between two types of gradients: first, we have the gradient of the neural network, $\nabla_\theta f(x; \theta)$, and then we have the gradient of the loss function, $\nabla_{\theta} \mathcal{L}(\theta)$, which is the one encountered more frequently in machine learning applications as this is used to minimize the loss.


With the preliminaries established, we can now describe the OGD algorithm in some detail. Please refer to the Appendix (\ref{OGD}) or the original paper for more information \citep{farajtabar2020orthogonal}.

    1. \textbf{Computing Model Gradients} After training the model on the first task in the usual manner, we first compute the model gradients of the class that corresponds to the ground truth label, that is, if an $x_i$ belongs to the $k$-th class, then we compute and store only $f_k(x_i; \theta)$. On the MNIST dataset, the authors recommend  computing the gradients at around 300 data points randomly sampled from the training set and thus the memory requirements are reasonable -- note that the dataset has about 60k training points.
    
    
    2.  \textbf{Orthogonal Basis} Once we have the model gradients $S$ = \{$\nabla_\theta f_k(x_i; \theta)$\}, we then compute the orthogonal basis for $S$ using the Gram-Schmidt method on all gradients. From here on, we take $S$ to denote the orthonormal set.
    
    3. \textbf{Computing Loss Gradients} Third, we compute the gradients of the loss function $\mathcal{L}$ on the data points that belong to the second task $\mathcal{T}_2$:
    \begin{align*}
        \vu \leftarrow \text{Stochastic/Batch Gradient for $\mathcal{T}_2$}.
    \end{align*}
    
    4. \textbf{Projection} We then project the gradient of the loss function computed in the previous step onto the subspace that is orthogonal to the span of the model gradients    \begin{align*}
        \vu' = \vu - \sum_{v \in S} \text{proj}_v(\vu).
    \end{align*}
    5. \textbf{Parameter Update} Finally, we update the parameters of the model using the new projected vector:
    \begin{align*}
        \theta \leftarrow \theta - \eta \vu',
    \end{align*}
    where we set $\eta$ is the step-size. In our experiments we use $\eta=0.001$.
     
     We repeat steps 3-5 until the model converges. In this report, we make slight modifications to the original approach and the modified approach is described in later sections.

\section{Related Work}

The problems discussed in this report are related to the ones explored in the field of machine unlearning \citep{cao2015towards}. This nascent field has been motivated in part by the recently introduced privacy regulations such as the General Data Protection Regulation in the European Union\citep{mantelero2013eu}, the California Consumer Privacy Act in the United States, and  Consumer Privacy Protection Act in Canada. While there are many popular approaches to data deletion \citep{bourtoule2021machine, ginart2019making, guo2019certified, garg2020formalizing, sekhari2021remember, du2019lifelong}, our approach is, at a high level, closely related to the one presented in \citep{golatkar2020eternal}; their method is based on a continual learning algorithm called Elastic Weight Consolidation \citep{kirkpatrick2017overcoming}, which uses the Fisher Information Matrix \citep{grosse2016kronecker} to assign importance to the learned weights. 

\section{Experiments}
In the following section we employ the MNIST dataset to perform two simple experiments to demonstrate unlearning and relearning. Unless otherwise stated, we use the standard train/test split of 60k/10k. Throughout, we use a simple convolutional neural network architecture that has around 20k parameters, which, when trained on 60k images, attains an accuracy of 98\% on the test set. We do not worry about tuning hyperparameters.


\subsection{Unlearning}\label{sec:unlearning}

\paragraph{Setup} We first look at the question of \textit{unlearning}. In order to do so, we take the MNIST dataset, and we purposefully mislabel two classes in the training dataset to create $(X,Y_{mis})$, i.e., it is the same as the original dataset but with two classes interchanged. Note that we leave the test set unchanged. For this experiment, we labelled the images containing the digit `2' as three and the images containing the digit `3' as two. We then train the CNN network network on the mislabelled training set, and it achieves an accuracy of 78\% on the untouched test set $(\tilde{X}, \tilde{Y})$, which is understandable given that the model has learned to mislabel some of the images.

\paragraph{Method} At the highest level, we approach this problem of debugging by exploiting the catastrophic forgetfulness of neural networks. That is, if we can somehow preserve the right kind of knowledge that the neural network has acquired, forgetting or unlearning the faulty behaviours comes naturally due to catastrophic forgetting. We do so with the help of modified OGD as follows:

     1. After training the model on the mislabelled training set, we compute the gradients of the model (not the loss) by randomly sampling a few hundred points from the training set without the mislabelled points. In other words, we sample from $(X, Y)$ with images corresponding to digits `2' and `3' removed. We then orthonormalize this set of gradient vectors (denoted by $S$). 

    2. Then, we randomly sample a vector that has as many elements as there are parameters in the model. That is, the random vector is of the same dimension as the gradient of the model (w.r.t the parameters). We choose to sample each element uniformly from the interval [0, 1]: $\vu \sim \text{uniform}(0, 1)$.
    
     3. Third, we project the randomly sampled vector onto the orthogonal subspace of the model gradients as follows: $\vu' = \vu - \sum_{v \in S} \text{proj}_v(\vu).$
     
     4. Finally, we update the parameters using the new projected vector: $\theta \leftarrow \theta - \eta \vu',$
    where we set $\eta =0.001$.
    
     5. We repeat steps 2-4 until the accuracy of the model on the (purposefully) mislabelled portion of the dataset is as good as random. We present the algorithm in the Appendix (\ref{unlearning}).

\begin{figure}
    \centering
    \includegraphics[scale=0.45]{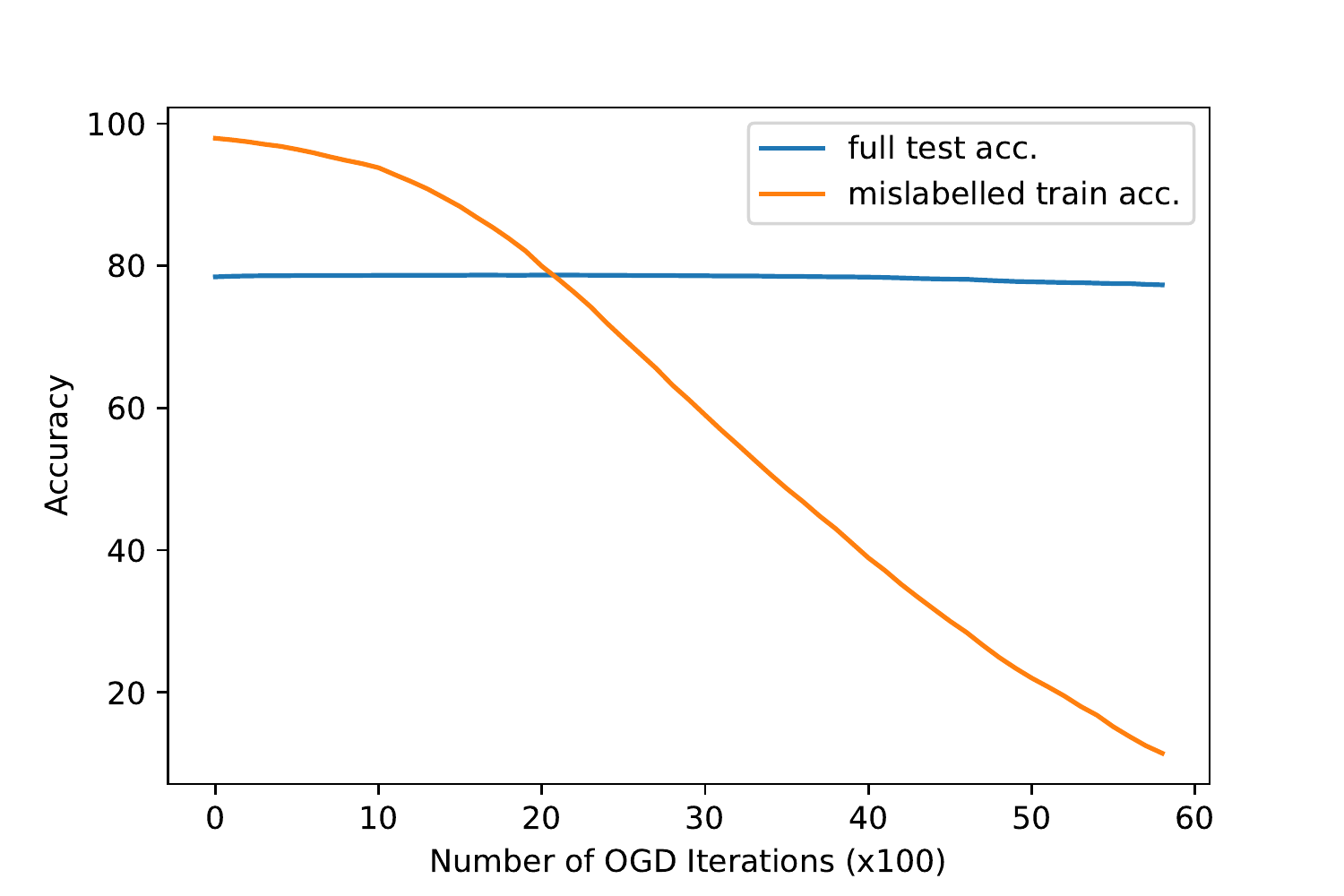}
    \includegraphics[scale=0.45]{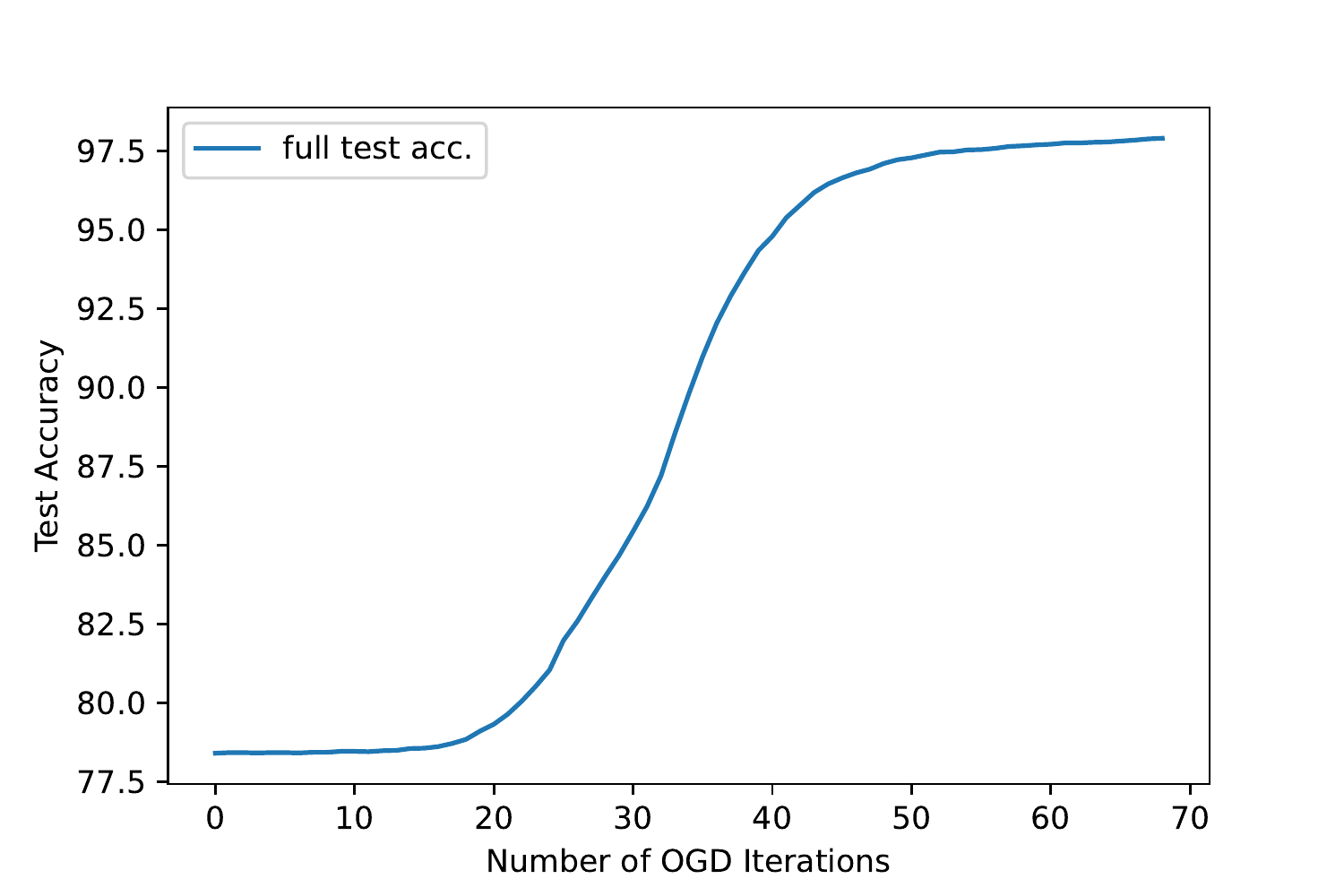}
    \caption{(left) Unlearning Experiment. The plot shows the accuracy of the model as a function of the number of OGD iterations (steps 2-4). The blue curve shows the accuracy of the model on the full test set $(\tilde{X}, \tilde{Y})$, and the orange curve shows the accuracy of the model on the purposefully mislabelled portion of the training dataset. (right) Relearning Experiment.  Here we plot the accuracy of the model on $(\tilde{X}, \tilde{Y})$ as a function of the number of OGD iterations.}
    \label{fig:unlearning}
\end{figure}

\paragraph{Results} On this simple example, we found that we can use modified OGD to unlearn the response of the model on on the misclassified data points, while preserving the general performance on everything else to a large extent. We see from Figure \ref{fig:unlearning} (left) that the model's accuracy on the purposefully mislabelled potion of the dataset right after training is close to perfect, but as we change the parameters of the model as detailed in steps 2-4, the model starts to unlearn this behaviour and instead responds to the input images of `2' and `3' in a random manner. We note that this unlearning generalizes to the test set as well. 

\subsection{Relearning}\label{sec:relearning}

\paragraph{Setup} Here the setup is similar to the unlearning case. We use the same CNN model and train it on the mislabelled training set $(X,Y_{mis})$, achieving around 78\% accuracy on the (untouched) test set.

\paragraph{Method} Here, in order for the model to relearn to respond in an appropriate way to all the inputs, including the initially mislabelled ones with minimal retraining, we do the following:

    1. Like before, after training the model on the mislabelled training set, we compute the gradients of the model (not the loss) by randomly sampling a few hundred points from the training set without the mislabelled points. In other words, we sample from $(X, Y)$ with images corresponding to digits `2' and `3' removed. We then orthonormalize this set of gradient vectors (denoted by $S$). 
    
    2. Then, we compute the gradient of the loss function using batches from a small portion of the training set that contains all the images of digits `2' and `3', with the right labels.
    
    3. Third, we project the gradient of the loss function computed in step 2 onto the orthogonal subspace of the model gradients as follows:
    \begin{align*}
        \vu' = \vu - \sum_{v \in S} \text{proj}_v(\vu).
    \end{align*}
    4. Finally, we update the parameters using the new projected vector:
    \begin{align*}
        \theta \leftarrow \theta - \eta \vu',
    \end{align*}
    where we set $\eta =0.001$.
    
    5. We repeat steps 2-4 until the accuracy of the model on the test set, $(\tilde{X}, \tilde{Y})$, reaches 98\%.
Please see the Appendix (\ref{relearning}) for the complete algorithm.

\paragraph{Results} The results are presented in Figure \ref{fig:unlearning} (right). We observe that, using the model gradients and the (corrected) mislabelled subset, we can train the model to an accuracy of 98\%, which is what the model attains when trained on the whole clean training set from scratch. We also note that using a simpler approach where we add an L2 term such as $\lambda || \theta^* - \theta ||^2$ (where $\theta^*$ denotes the weights of the model after training on task 1) to the loss fucntion while training on the second task does not work.

\section{Conclusion}
We present two experiments supporting the hypothesis that the problem of debugging can be treated as a continual learning problem. First, we use the phenomenon of catastrophic forgetting to our advantage, and show that we can in fact do targeted unlearning. The model forgets to (mis)-classify the images containing the digits `2' as three and vice-versa. However, when we compare the average (highest) confidence with which the model classifies these images before and after the unlearning, we did not notice a drastic difference: 0.98 vs 0.78 -- when the weights are randomly initialized the confidence is around 0.12. In the ideal case, not only would the model unlearn, but the confidence of the predictions would also look more like the randomly initialized model. Second, we present an experiment on relearning and demonstrate that the model can successfully relearn the right behaviour, reaching the maximum possible accuracy on the dataset, given the model architecture. 

We believe that it would be interesting to apply these methods to deal with problems where the model has a lot more parameters and/or where the dataset is much larger. We have also considered the possibility of using modified OGD (like in the unlearning case) to increase the robustness of models against targeted adversarial attacks.

\clearpage

\bibliography{main}
\bibliographystyle{abbrv}

\clearpage
\appendix

\begin{algorithm}[t]
\label{OGD}
\caption{OGD}
\KwIn {Task sequence $T_A$} 

\While{stopping criterion not met}{

$g \leftarrow \nabla_{\theta} \sum_i \mathcal{L}(f(x^A_i; \theta), y^A_i)$

$ \theta \leftarrow \theta -\eta g$
}
\bigskip
Sample $m$ examples from $T_A$: $t^A$ = $\{(x^A_1, y^A_1), \hdots, (x^A_m, y^A_m) \}$\\
\textbf{Initialize} $S = \{\}$\\
\For{$(x,y) \in t^A$}{

$u=\nabla_{\theta} f(x;\theta^A)-\sum^{}_{v\in S} \text{proj}_{v} (\nabla_{\theta} f(x;\theta^A))$

$S=S\bigcup{u}$
}
\bigskip
\KwIn {Task sequence $T_B$} 
\While{stopping criterion not met}{

$g \leftarrow \nabla_{\theta} \sum_i \mathcal{L}(f(x^B_i; \theta), y^B_i)$\\
$\tilde{g}=g-\sum^{}_{v\in S} \text{proj}_{v} (g)$\\
$ \theta \leftarrow \theta -\eta \tilde{g}$
}
\end{algorithm}

\begin{algorithm}[t]
\label{unlearning}
\caption{Modified OGD for Unlearning}
\KwIn {Task sequence $T_A$: Mislabelled MNIST} 

\While{stopping criterion not met}{

$g \leftarrow \nabla_{\theta} \sum_i \mathcal{L}(f(x^A_i; \theta), y^A_i)$

$ \theta \leftarrow \theta -\eta g$
}
\bigskip
$t^A:$ sample from $\{T_A - $ mislabelled examples\}\\
\textbf{Initialize} $S = \{\}$\\
\For{$(x,y) \in t^A$}{

$u=\nabla_{\theta} f(x;\theta^A)-\sum^{}_{v\in S} \text{proj}_{v} (\nabla_{\theta} f(x;\theta^A))$

$S=S\bigcup{u}$
}
\bigskip
\While{stopping criterion not met}{

$g \sim \text{uniform}[0, 1]$\\
$\tilde{g}=g-\sum^{}_{v\in S} \text{proj}_{v} (g)$\\
$ \theta \leftarrow \theta -\eta \tilde{g}$
}
\end{algorithm}

\begin{algorithm}[t]
\label{relearning}
\caption{Mofified OGD for Relearning}
\KwIn {Task sequence $T_A$: Mislabelled MNIST} 

\While{stopping criterion not met}{

$g \leftarrow \nabla_{\theta} \sum_i \mathcal{L}(f(x^A_i; \theta), y^A_i)$

$ \theta \leftarrow \theta -\eta g$
}
\bigskip
$t^A:$ sample from $\{T_A - $ mislabelled examples\}\\
\textbf{Initialize} $S = \{\}$\\
\For{$(x,y) \in t^A$}{

$u=\nabla_{\theta} f(x;\theta^A)-\sum^{}_{v\in S} \text{proj}_{v} (\nabla_{\theta} f(x;\theta^A))$

$S=S\bigcup{u}$
}
\bigskip
\KwIn {Task sequence $T_B$: data with desired labels} 
\While{stopping criterion not met}{

$g \leftarrow \nabla_{\theta} \sum_i \mathcal{L}(f(x^B_i; \theta), y^B_i)$\\
$\tilde{g}=g-\sum^{}_{v\in S} \text{proj}_{v} (g)$\\
$ \theta \leftarrow \theta -\eta \tilde{g}$
}
\end{algorithm}

\end{document}

%% file: math.tex
\usepackage{amsmath,amsfonts,bm}

\def\eqref#1{equation~\ref{#1}}

\def\1{\bm{1}}

\def\vu{{\bm{u}}}

\DeclareMathAlphabet{\mathsfit}{\encodingdefault}{\sfdefault}{m}{sl}
\SetMathAlphabet{\mathsfit}{bold}{\encodingdefault}{\sfdefault}{bx}{n}

\newcommand{\R}{\mathbb{R}}

